\documentclass{article}

\usepackage{arxiv}

\usepackage[T1]{fontenc}
\usepackage[utf8]{inputenc}
\usepackage{hyperref}
\usepackage{url}
\usepackage{booktabs} 
\usepackage{amsfonts}
\usepackage{amsthm}
\usepackage{amsmath}
\usepackage{nicefrac}
\usepackage{microtype}
\usepackage{cleveref}
\usepackage{graphicx}
\usepackage[numbers, compress]{natbib}
\usepackage{doi}
\usepackage{siunitx}
\usepackage{algorithmic}
\usepackage{textcomp}
\usepackage{xcolor}
\usepackage{fancyhdr}
\usepackage{hyperref}
\usepackage{soul}
\usepackage{subcaption}
\usepackage{algorithm}
\usepackage{multirow}
\usepackage{makecell}

\title{DeepHYDRA: Resource-Efficient Time-Series Anomaly Detection in Dynamically-Configured Systems}

\usepackage{authblk}

\newbox{\orcid}\sbox{\orcid}{\includegraphics[scale=0.06]{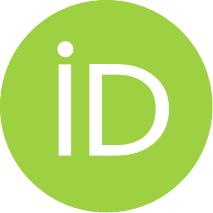}} 
\author[1,2]{%
	\href{https://orcid.org/0000-0002-0426-5837}{\usebox{\orcid}\hspace{1mm}Franz Kevin Stehle\thanks{\texttt{kevin.stehle@ziti.uni-heidelberg.de}}}%
}
\author[2]{%
	\href{https://orcid.org/0000-0001-6581-9410}{\usebox{\orcid}\hspace{1mm}Wainer Vandelli\thanks{\texttt{wainer.vandelli@cern.ch}}}%
}
\author[2]{%
	\href{https://orcid.org/0000-0003-2664-3437}{\usebox{\orcid}\hspace{1mm}Giuseppe Avolio\thanks{\texttt{giuseppe.avolio@cern.ch}}}%
}
\author[2]{%
	\href{https://orcid.org/0000-0003-3796-3620}{\usebox{\orcid}\hspace{1mm}Felix Zahn\thanks{\texttt{felix.zahn@cern.ch}}}%
}
\author[1]{%
	\href{https://orcid.org/0000-0001-9562-0680}{\usebox{\orcid}\hspace{1mm}Holger Fröning\thanks{\texttt{holger.froening@ziti.uni-heidelberg.de}}}%
}
\affil[1]{Computing Systems Group, Institute of Computer Engineering, Heidelberg University, Germany}
\affil[2]{ATLAS Experiment, CERN, Geneva, Switzerland}

\renewcommand{\shorttitle}{\textit{Stehle et al.:} DeepHYDRA}

\hypersetup{
	pdftitle={DeepHYDRA: Resource-Efficient Time-Series Anomaly Detection in Dynamically-Configured Systems},
	pdfsubject={cs.AI},
	pdfauthor={Franz Kevin Stehle, Wainer Vandelli, Giuseppe Avolio, Felix Zahn, Holger Fröning},
	pdfkeywords={Anomaly Detection, Clustering, DBSCAN, Deep Learning, Transformers, Dimensionality Reduction},
}

\begin{document}
\maketitle

\begin{abstract}
	Anomaly detection in distributed systems such as High-Performance Computing (HPC) clusters is vital for early fault detection, performance optimisation, security monitoring, reliability in general but also operational insights.
	It enables proactive measures to address issues, ensuring system reliability, resource efficiency, and protection against potential threats.
	Deep Neural Networks have seen successful use in detecting long-term anomalies in multidimensional data, originating for instance from industrial or medical systems, or weather prediction.
	A downside of such methods is that they require a static input size, or lose data through cropping, sampling, or other dimensionality reduction methods, making deployment on systems with variability on monitored data channels, such as computing clusters difficult.
	To address these problems, we present DeepHYDRA (Deep Hybrid DBSCAN/Reduction-Based Anomaly Detection) which combines DBSCAN and learning-based anomaly detection.
	DBSCAN clustering is used to find point anomalies in time-series data, mitigating the risk of missing outliers through loss of information when reducing input data to a fixed number of channels.
	A deep learning-based time-series anomaly detection method is then applied to the reduced data in order to identify long-term outliers.
	This hybrid approach reduces the chances of missing anomalies that might be made indistinguishable from normal data by the reduction process, and likewise enables the algorithm to be scalable and tolerate partial system failures while retaining its detection capabilities.
	Using a subset of the well-known SMD dataset family, a modified variant of the Eclipse dataset, as well as an in-house dataset with a large variability in active data channels, made publicly available with this work, we furthermore analyse computational intensity, memory footprint, and activation counts.
	DeepHYDRA is shown to reliably detect different types of anomalies in both large and complex datasets.
	At the same time, the applied reduction approach is shown to enable real-time anomaly detection of a whole computing cluster while occupying proportionally minuscule compute resources, enabling its usage on existing systems without the need for hardware changes.
\end{abstract}

\begin{paragraph}
\textcopyright {Franz Kevin Stehle | ACM} {2024}. This is the author's version of the work. It is posted here for your personal use. Not for redistribution. The definitive Version of Record was published in {Proceedings of the 38th ACM International Conference on Supercomputing (ICS '24), June 4--7, 2024, Kyoto, Japan}, \hyperlink{https://dx.doi.org/10.1145/3650200.3656637}{https://dx.doi.org/10.1145/3650200.3656637}.
\end{paragraph}

\section{Introduction}
The recent years have seen an enormous increase in the need for high-dimensional time-series data analysis as well as methods to serve those needs, from computing-cluster-based processing~\cite{gomes2019survey,almeida2023time} on one side of the spectrum to edge computing~\cite{oyekanlu2017predictive,park2018lired} applications on the other.
Traditional time-series analysis methods often times perform poorly on the high-dimensional data found in such applications~\cite{nambiar2020transforming}, and thus the field of time-series data analysis has seen an explosion of deep-learning-based methods that intend to overcome these issues~\cite{zhou2021informer}.
The field of time-series anomaly and novelty detection is no exception to this trend~\cite{doshi2022tisat,DBLP:journals/corr/abs-2201-07284}.\newline
These deep-learning methods require input data to be of constant dimensionality throughout training and deployment.
While this usually does not pose a problem, it does when observing dynamically-configured systems.
Examples of this are high-performance computing (HPC) clusters, where failures resulting in lengthy repair times are somewhat common~\cite{schroeder2009large}, in turn resulting in variability in cluster configuration, as well as highly-modular IoT applications~\cite{benammar2018modular,codeluppi2020lorafarm}.
This poses a challenge for fixed-dimensional input deep-learning methods, as the variable-sized data coming from such systems has to be brought to a fixed size to allow processing.
While it is possible to achieve this through a range of established reduction methods, the reduction of time-series data in a lossy manner may hide anomalies that only impact few of the observed data channels.\newline
\begin{figure}[t]
	\centering
	\includegraphics[alt={A selection of data channels from different sources are reduced to their median and standard deviation. The unreduced data is processed by the DBSCAN anomaly detector, while the reduced data is processed by a long-term anomaly detection method. The predictions of both methods are combined to a unified anomaly score.},width=0.65\columnwidth]{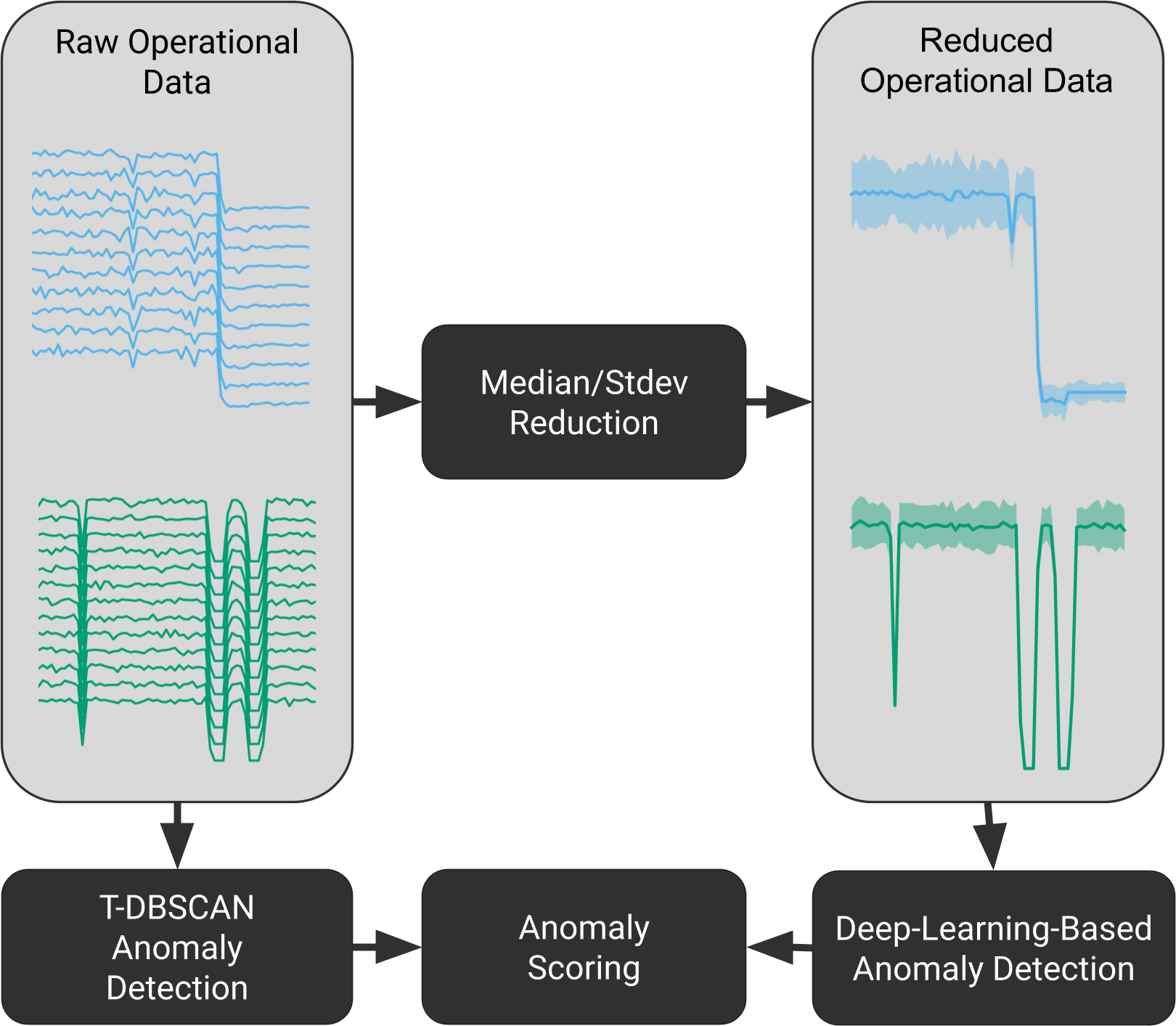}
	\caption{Schematic of the dataflow in the proposed DeepHYDRA anomaly detection method.}
	\label{fig:offline-detection-pipeline}
\end{figure}
The High-Level Trigger (HLT) system of the ATLAS particle detection experiment at CERN~\cite{atlas2016atlas} is a prime example of a dynamically-configured system.
This heterogeneous system consisting of commercial-off-the-shelf (COTS) servers performs highly-parallel data-processing in a loosely-coupled manner and serves as the primary target for the methods presented in this work.
It is composed of racks of different generations of hardware, resulting in homogeneous subgroups inside the heterogeneous system.
As it is required to be highly-available during physics data-taking, maintenance may be performed while the system is online, resulting in a change in dimensionality in the observed data.
At the same time, the enormous energy costs required in operating the Large Hadron Collider (LHC)~\cite{cern2022powering} make it necessary to conduct much of the component testing before the system is fully operational, resulting in a shift between the testing and operation environment.
This increases the chances of encountering anomalies in the operational stage, which are likely to be of unpredictable nature.\newline
The aforementioned variability in features that require monitoring inherent in the HLT combined with its high availability requirements result in the need for an anomaly detection method that is both reliable and scalable in the number of monitored data channels.
To address this challenge, we propose a parallel approach that leverages the strengths of two distinct anomaly detection methods, which we name \textbf{D}eep \textbf{Hy}brid \textbf{D}BSCAN/\textbf{R}eduction-Based\textbf{A}nomaly {D}etection, or \textit{DeepHYDRA} for short.
The approach is enabled through massive dimensionality reduction combined with the parallel application of a fast detection method on the unreduced data aimed at mitigating a loss in recall through loss of information.
This strategy is based on a variation of the DBSCAN clustering algorithm~\cite{ester1996density} that enables reliable anomaly detection in high-dimensional and variable-dimensional data streams by taking advantage of existing homogeneity in the monitored heterogeneous system.
The second part of the proposed strategy consists of a transformer-based anomaly detection method applied to the reduced data stream.
This two-fold detection strategy is visualised in Figure~\ref{fig:offline-detection-pipeline}.
For this purpose, a selection of existing transformer models intended for anomaly detection, including a novel, semi-supervised approach are evaluated.\newline
Furthermore, the reduction methodology introduced in this work not only has the effect of making observation of dynamically configured systems possible, it also potentially offers a massive reduction in terms of computational intensity and memory requirements.
This opens the door to use the proposed data pipeline not only in systems that are dynamically configured, but also have tight real-time requirements or scarce computational resources, for example miniaturised sensor networks~\cite{iyer2022wind}.
We therefore also quantify the computational intensity and size requirements of the models discussed in this work in order to gain insights into the effects of  the presented combination of DBSCAN with long-term anomaly detection machine learning models.
The key contributions made as part of this work are as follows:
\begin{enumerate}
	\item A method for efficient deployment of neural-network-based anomaly detection to detect long-term anomalies in high-dimensional and variable-dimensional data streams is proposed.
	\item A time-series anomaly detection training regimen is presented that includes time-domain and frequency-domain data augmentation and semi-supervised training.
	\item We present a high-dimensional time-series anomaly detection dataset based on historic HLT operational data, which is injected with synthetic anomalies, and is made public as part of this work\footnote{The datasets are available under \url{https://zenodo.org/record/7908064}. The code used in this work can be found under \url{https://github.com/UniHD-CEG/DeepHYDRA}}.
	\item The performance of the used transformer methods are first evaluated on the machine-1-1 subset of the SMD dataset~\cite{su2019robust}, followed by an evaluation on the introduced in-house dataset that includes analysis of the impact of semi-supervised training and data augmentation.
	An evaluation on a version of the Eclipse dataset presented by Aksar et al.~\cite{aksar2023dataset} adapted to the DeepHYDRA dataflow is conducted to test the generalisability of the proposed approach to HPC systems aside from the HLT.
	The combined detection strategy is evaluated on the best-performing candidates.
	\item Lastly, the methods considered in this work are compared in terms of inference FLOPs and memory requirements.
\end{enumerate}
A discussion of the advantages as well as the shortfalls and limitations of this method and possible avenues of development concludes this work.

\section{Background and Related Work}
In this section, we give an overview of the different methods used as part of DeepHYDRA, as well as other approaches followed in the respective field.
This serves to put the components used as part of DeepHYDRA into context, as well as to lay out where our approach differs from the existing literature.

\subsection{Anomaly Detection in HPC Clusters}
Many different methods have been proposed for anomaly detection in HPC clusters.
Molan et al. propose the Recurrent Unsupervised Anomaly Detection (RUAD) method for unsupervised anomaly detection in HPC clusters~\cite{molan2023ruad}.
This approach uses an autoencoder-based detection approach using an LSTM-encoder to capture temporal dependencies, and then uses the autoencoder reconstruction error as an anomaly measure.
This approach is comparable to the transformer-based detection data path used in DeepHYDRA, and also analyses system metadata, such as core temperatures, CPU usage, and memory usage.
The dataset of study for RUAD is much lower-dimensional, and as such the detection pipeline does not include dimensionality reduction aside from the reduction from the input space to the latent space inherent in autoencoders.
Unlike DeepHYDRA, which requires anomaly-free training data for unsupervised training, RUAD can tolerate a low amount of anomalies in the training data.
DeepHYDRA on the other hand is able to leverage anomalies in labeled training data using semi-supervised training.\newline
Another recent approach based on autoencoder reconstruction error is presented by Borghesi et al.~\cite{borghesi2019online}.
Unlike RUAD and DeepHYDRA, where centrally collected metrics are monitored~\cite{molan2023ruad}, this method is deployed to the edge of each compute node separately.
\subsection{Dimensionality Reduction}
A multitude of dimensionality reduction methods have been employed for tabular and time-series data.
One of the most widely-used and well-known of such methods is Principal Component Analysis (PCA)~\cite{jolliffe2016principal,vasan2016dimensionality,salem2019data}.
This linear and unsupervised dimensionality reduction method computes the principal components of a given input dataset, linear combinations that capture the greatest variance in the data~\cite{jolliffe2002springer}.
It has been adapted to many different data types~\cite{jolliffe2016principal}, with time-series data being one of them~\cite{kane2017multivariate}.
The extracted principal components may improve the performance of subsequent algorithms by capturing important features of the input data~\cite{patil2022network}.\newline
Among neural-network-based dimensionality reduction methods, autoencoders are a popular option for tabular data~\cite{ruff2019deep} and time-series data~\cite{sakurada2014anomaly}.
Compared to these sophisticated dimensionality reduction methods, the method followed for DeepHYDRA is a more simplistic median- and standard-deviation-reduction that is tailored to the cluster structure of the observed system.
In this it is restricted in its possible applications, but in turn provides better interpretability of any anomaly detection performed on the reduced data.
Anomalies detected in the principal components extracted in PCA or the latent space vectors produced by autoencoders cannot easily be traced back to an individual data subgroup, e.g. a server rack.
The per-subgroup reduction approach followed in DeepHYDRA on the other hand produces reduced data from which the source of a given failure is readily apparent, thus improving anomaly prediction interpretability.

\subsection{Clustering in Anomaly Detection}
Clustering is a popular approach for unsupervised anomaly detection in fields such as fraud detection~\cite{dharwa2011data} and intrusion detection~\cite{prodanoff2020anomaly}.
Clustering methods often applied in anomaly detection include  the classical clustering methods of k-means clustering~\cite{jain2010data} and  DBSCAN~\cite{ester1996density}.
These methods famously suffer from the "curse of dimensionality" when applied to high-dimensional datasets~\cite{kriegel2009clustering}.
This stems from, among other things, irrelevant features present in a given high-dimensional dataset, and the existence of correlations between attributes only relevant for some clusters, but not for others.\newline
For the DBSCAN-based anomaly detection employed in this work this does not pose a problem, as we do not cluster the dataset as a whole, but each input sample individually, treating each data channel as an individual data point instead of a dimension in the input data.
The combination of DBSCAN-based anomaly detection in combination with transformer-based detection is at the time of writing not yet found in existing literature to the best of our knowledge, nor is our approach of its per-sample application for variable-size data.
\subsection{Deep Learning in Time-Series Anomaly Detection}
Many different deep learning methods have been applied to the field of anomaly detection.
We test a varied cross section of models for usage as part of our proposed DeepHYDRA approach, which are described in the following.\newline
Zong et al. introduce the Deep Autoencoding Gaussian Mixture Model (DAGMM)~\cite{zong2018deep}, which extends the established Gaussian Mixture Model methodology to higher-dimensional datasets by combining it with an autoencoder stage for dimensionality reduction.
OmniAnomaly is a Bayesian method employing GRU layers to capture long-term dependencies in multi-variate data~\cite{su2019robust}.
The USAD method proposed by Audibert et al. is based on a variation of the known autoencoder architecture, extending it with a second decoder~\cite{audibert2020usad}.
The decoders are trained in an adversarial fashion using a two-step training methodology, with the second decoder learning to discriminate between reconstructions of the first fed back into the encoder.
During inference, the anomaly scoring is then derived from the outputs of both decoder stages of the model.
Also included in our investigation are transformers, which have been shown to deliver competitive performance in the task of long-term time-series prediction~\cite{zhou2021informer} as well as time-series anomaly detection~\cite{doshi2022tisat}.
One such method is TranAD, proposed by Tuli et al.~\cite{DBLP:journals/corr/abs-2201-07284}, employing two transformer encoders and decoders each that are trained in an adversarial fashion.
Another transformer-based anomaly detection solution is TiSAT, introduced by Doshi et al.~\cite{doshi2022tisat}.
This model combines an Informer for prediction~\cite{zhou2021informer} with a k-Nearest Neighbor (kNN) model, with anomaly scoring based on the distance of a given prediction in the kNN model relative to predictions encountered during training.
The informer-based detection methods employed in this work are based on the work Doshi et al.~\cite{doshi2022tisat}, but leave out the kNN part of that model, as it has shown subpar results in early testing.
Notably, all of the discussed models are unsupervised in the sense that they are intended to be trained on unlabelled, known-good data.\newline
The Prodigy framework proposed by Aksar et al.~\cite{aksar2023prodigy} is a deep anomaly detection method tailored to HPC workloads.
Unlike the other methods described thus far, Prodigy is intended for post-hoc anomaly detection on a number of features extracted from performance monitoring traces using the TSFRESH toolkit~\cite{christ2018time}.
It is based on the well-known autoencoder reconstruction error-based detection approach.
What sets Prodigy apart from other methods is that the anomaly detector only represents a part of the framework, with other functions being included being generic data generation, preprocessing, and feature extraction.
This makes the Prodigy approach overall simple to set up and integrate into existing monitoring pipelines.

\subsection{Time-Series Data Augmentation}
While data augmentation is commonplace in deep learning domains such as image classification~\cite{shorten2019survey}, comparatively little research has been published about data augmentation for time-series, and even less for time-series anomaly detection \cite{wen2021time}.
Gao et al. show that the RobustTAD model they propose can achieve a  4.72 \% higher F1 score compared to the same model trained on unaugmented data~\cite{gao2020robusttad}.
The applied augmentations in this case are flipping, cropping, label expansion, and frequency space-based perturbations of the amplitude and phase.
The latter method is named amplitude and phase perturbations (APP) by Wen et al.~\cite{wen2021time}, and we will continue to refer to such methods as such in the remainder of this publication.
Another data augmentation method named Iterated Amplitude Adjusted Fourier Transform (IAAFT)~\cite{venema2006stochastic}, creating surrogate data of comparable power spectrum to the original data, has been employed by Lee et al.~\cite{lee2021studies} to increase the performance of an image classification-based time-series classification model.
This is achieved by adding augmented data to the training set of up to 100 times the size of the original training set, while simultaneously adjusting for class imbalances in the original set.\newline
A comparison of different augmentation methods is not within the scope of this work, and thus we base our augmentation strategies solely on the work by Wen et al.~\cite{wen2021time}, as it presents multiple candidate methods of varied nature.

\section{Method}
The combined detection method aims to both identify long-term anomalies influencing the system as a whole as well as short-term anomalies only influencing some elements of the system. For this purpose, we devise a hybrid clustering and deep-learning-based approach that is detailed in the following.
\subsection{DBSCAN-Based Detection}
\begin{figure}
	\centering
	\includegraphics[alt={While in the upper schematic, all the data from a given subgroup is identified by the clustering algorithm to belong to the same cluster, in the bottom schematic, data from two of the channels is identified as belonging to different clusters. This change in cluster membership is used as a anomaly detection method in the T-DBSCAN approach.},width=0.4\columnwidth]{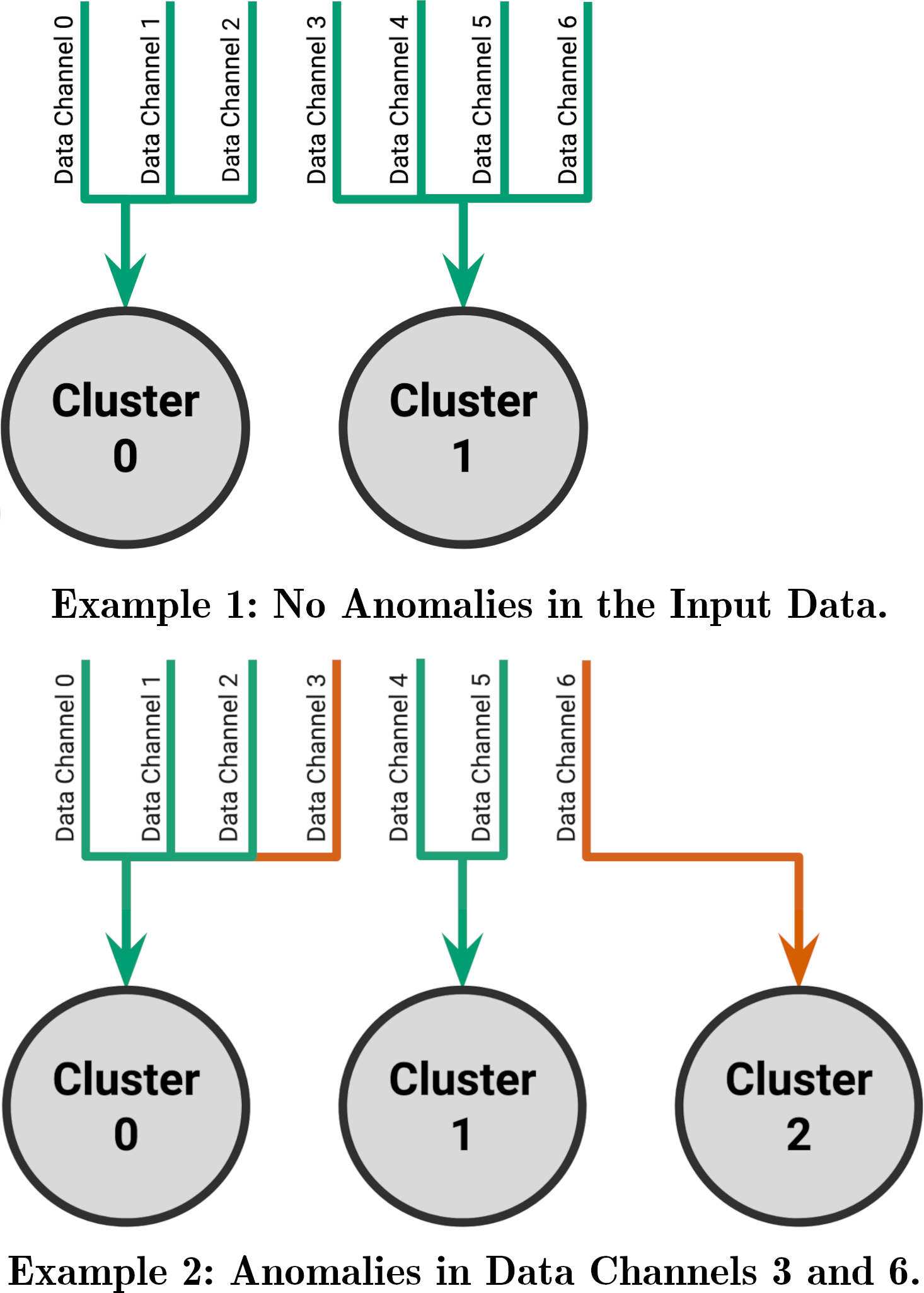}
	\caption{T-DBSCAN-based approach to anomaly detection using prior knowledge of system properties. In these examples, the data channels 0, 1, and 2, form a known subgroup, as do channels 3 through 6. In Example 1, the data clustering is identical to the known subgroup membership. In Example 2, anomalies are detected in Data Channels 3 and 6, as the cluster membership of the produced data deviates from the other channels of the known subgroup.}
	\label{fig:dbscan-clustering-anomaly-detection}
\end{figure}

Many heterogeneous distributed computing applications still retain some homogeneity on a local level.
This can take the form of spatially local homogeneity arising from hierarchical interconnection speeds between processing nodes.
For example, the communication bandwidths and latencies between the cores in a processor, between NUMA nodes, or between nodes in the same server rack are starkly different compared to one another, but identical inside of each locality level.
At the same time, processing node hardware configuration in a heterogeneous system may also be locally homogeneous.
Reasons for such a configuration may be maintainability or cost reduction during system upgrades.\newline
Such local system homogeneity often times also manifests in local homogeneity in the behaviour of processes running on a distributed system. If this is the case, it may be leveraged to detect anomalies in individual processing nodes of the system.
The ATLAS HLT is a system where such local homogeneity manifests in a multitude of operational data, arising from non-uniform, but locally homogeneous hardware configurations.
Clustering is an obvious choice for detecting short-term anomalies between systems of known locally correlated behaviour.
DBSCAN, a popular choice for clustering-based anomaly detection~\cite{loi2020abnormal}, offers the benefit of not having to provide prior knowledge of the number of clusters in a given dataset.
The DBSCAN implementation used in this work is the implementation provided by the Scikit-Learn framework~\cite{scikit-learn}.
This is central to the proposed approach, as it is based on the assumption that for normal data, the clustering algorithm clusters data pertaining to the channels of a subgroup into a single cluster, while it assigns outlier data into different clusters.
This approach combines the inherent ability of DBSCAN to label noisy samples with the prior knowledge of intended system behaviour.
This allows the detection of not only noise-like outliers with no inherent correlation between them, but also the detection of outlier data channels that are tightly bound together, forming a cluster of their own.\newline
As opposed to applying DBSCAN to the dataset as a whole, our method applies DBSCAN to individual per-timestep data slices, effectively treating the data as 1-dimensional with a length equal to the number of channels.
In addition to the data themselves, the intended subgroup identifier is provided as a-priori knowledge to improve detection performance.
We do not retain any information between data slices, effectively retraining DBSCAN on each individual slice, which only has negligible computational cost because of the low effective dimensionality of the data.
The cluster memberships themselves are then compared to the previous timestep, and a state machine logs channels that have changed membership.
This application of DBSCAN on transposed input data will be called T-DBSCAN in the remainder of this publication.\newline
The functioning of this approach when encountering normal and anomalous data is visualised in Figure~\ref{fig:dbscan-clustering-anomaly-detection}.
The specific approach followed for the HLT dataset requires the a priori knowledge of the actual subgroup a data channel is a member of, though such memberships could be obtained from training data in applications where such data is not readily available.
\subsection{Dimensionality Reduction}
Like most time-series anomaly detection models, the deep-learning-based models analysed in this work require fixed input data dimensionality.
As this cannot be guaranteed in the given application scenario, a method to reduce the variable size input to a fixed format is needed.
For this purpose, the data of individual channels in a subgroup is reduced to the per-subgroup median and standard deviation of the data, serving as representatives of a subgroup as a whole.
This approach allows the combined anomaly detection pipeline to tolerate a large number of failing data channels, while still being able to keep performing anomaly detection on the active channels.
We decide against usage of the mean and standard deviation to represent a subgroup in favour of the median and standard deviation, as the purpose of the learning-based detection within DeepHYDRA is to detect long-term anomalies that influence the subgroup as a whole, while DBSCAN is employed with the intention of detecting anomalies within a subgroup.
Usage of the median in place of the mean makes the reduced data less dependent on outliers occurring within only a handful of channels within a subgroup.
\subsection{Learning-Based Detection}
Six deep-learning-based model candidates are evaluated to serve as the long-term anomaly detection methods for DeepHYDRA, with three coming from the category of transformers, and the remaining three representing a set of more well-established deep-learning-based approaches.
These three non-transformer candidates are OmniAnomaly~\cite{su2019robust}, DAGMM~\cite{zong2018deep}, and USAD~\cite{audibert2020usad}.
On the transformer side, the TranAD model~\cite{DBLP:journals/corr/abs-2201-07284} and two approaches based on the informer~\cite{zhou2021informer} model with anomaly scoring based on the Mean Squared Error (MSE) between prediction and actual data are considered.
This approach is inspired by code provided by Doshi et al.~\cite{doshi2022tisat}.
While the first informer-based method uses the MSE loss for both training and anomaly scoring, the third transformer candidate employs a novel semi-supervised training methodology described in the following.

\subsubsection{Semi-Supervised Training}
We introduce a MSE-based loss function allowing for semi-supervised training named Supervised MSE, or SMSE.
The goal of this loss function is to improve detection performance by training the model on a small subset of labelled data, in addition to the larger subset of unlabelled data.
This method is inspired by the Deep SAD model proposed by Ruff et al.~\cite{ruff2019deep}, which is able to achieve significant improvements in anomaly detection performance by introduction of a semi-supervised training methodology and a loss function to enable it, without changing other model parameters.\newline
$l_{\text{SMSE}}$, shown in Equation~\ref{eq:l-smse}, is identical to MSE loss on unlabelled data.
$X_i$ denotes the model input at index $i$, while $\hat{X_i}$ represents the reconstructed input.
On labelled data however, we instead penalise reconstruction of anomalies by inverting the reconstruction MSE.
The amount of penalization is dependent on the factor $\phi$, representing the ratio of anomalies in the input window of length $n$, and calculated as described in Equation~\ref{eq:phi}, where $\delta$ denotes the Kronecker Delta function.
The $\tanh$ function is applied to the inverted MSE loss in order to avoid exploding losses when encountering anomalous data.
Lastly, a scaling factor $\mu$ is applied to the semi-supervised part of the SMSE loss in order to scale the produced losses to the range encountered in unlabelled data.
\begin{align}
	\label{eq:l-smse}l_{\text{SMSE}} &= \!\begin{aligned}[t]&\frac{(1 - \phi)}{n}\sum_{i=1}^{n}(X_i - \hat{X_i})^2 + \\ &\mu\phi\tanh\left(\frac{n}{\sum_{i=1}^{n}(X_i - \hat{X_i})^2}\right) \end{aligned}\\
	\label{eq:phi}\phi &=  \frac{1}{n}\sum_{s \in S_1^n} \delta_{s,1}
\end{align}
The scaling with the number of anomalous instances in the input data aims at more harshly penalising long-term anomalies, while not reducing reconstruction performance on normal data in input windows with only few anomalies.
\subsection{Data Augmentation}
\label{subsec:data-augmentation}
\begin{figure}
	\centering
	\includegraphics[alt={The event rates, reaching around 60 Hz at their maximum, show a characteristic pattern of peaks, followed by a slow decrease down from that peak, followed by the next peak,with an overall decreasing trend. The data shows multiple short peaks down to zero. The figure shows multiple distinct bands that correspond to different hardware generations inside the HLT. The period depicted spans 16 hours.},width=0.55\columnwidth]{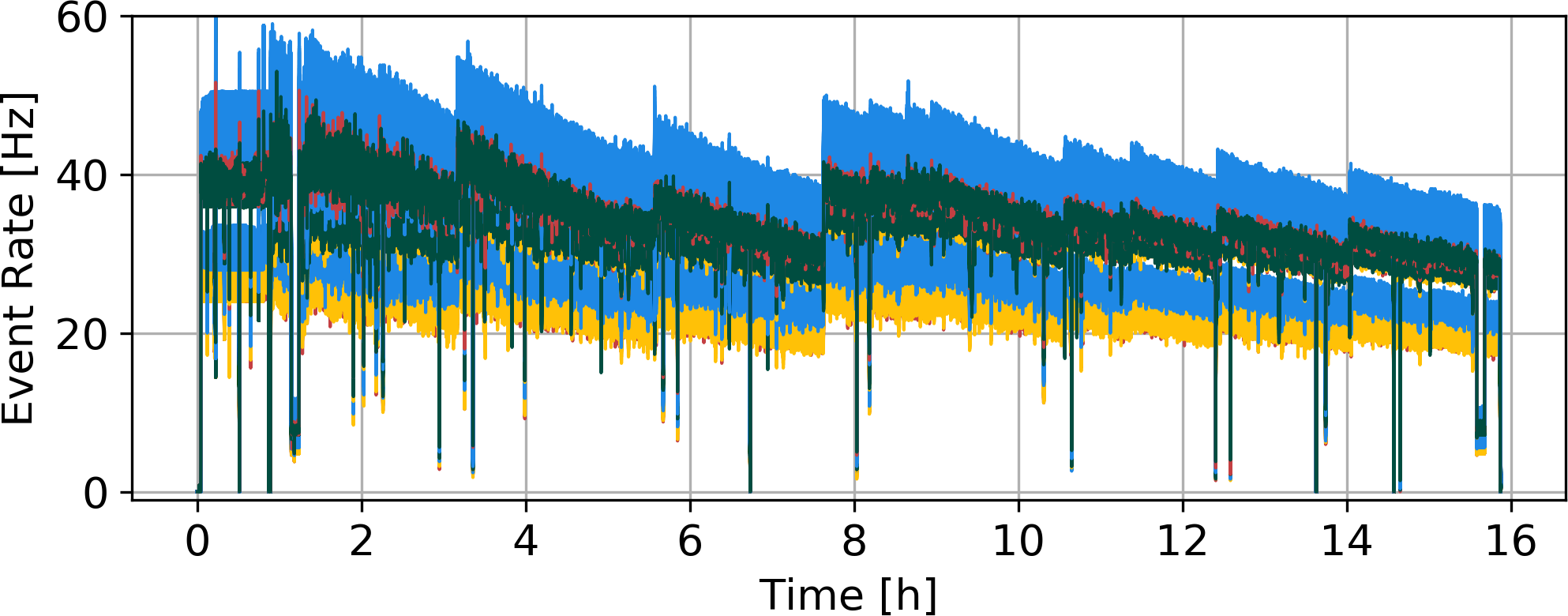}
	\caption{Example of a typical per-machine data event rate development throughout a physics data-taking session, with the individual channels coloured by processor type. The update interval is 5 s.}
	\label{fig:DCM-rate-example}
\end{figure}
The novel datasets presented in this work consist of per-machine data event rates of the nodes of the HLT.
These rates are an important metric in judging the overall system state, and many failure modes of lower-level components eventually produce observable anomalies in this data.
As such, they are a prime target of interest for the application of the DeepHYDRA approach.\newline
These data rates show a characteristic progression throughout a session of physics data taking, as shown in Figure~\ref{fig:DCM-rate-example}.
Although there exists an immense amount of recorded data, at the same time, it takes expert knowledge to discern any anomalies in existing data, and even a skilled expert might miss anomalies in the thousands of data points collected during a session.
While most sessions follow a characteristic shape, they do vary significantly in absolute range, depending on the currently performed physics experiment.\newline
Two different augmentation strategies are employed to increase the detection performance based on training with a limited amount of known anomaly-free data:
one strategy is the simple scaling of the data within a given range by a random factor.
The second strategy employed is the before mentioned scaling, combined with Amplitude and Phase Perturbation (APP), as described by Wen et al.~\cite{wen2021time}.
The composed training set data is split into segments containing individual sessions, and either scaling, scaling and APP, or no augmentation is applied to them, depending on the desired ratio of augmented data to original data in the resulting dataset.
The scaling and APP-based augmentation approach introduces significant invariances in amplitude scaling and noise into the training data while retaining the characteristic development the analysed data shows during physics data taking.\newline
For the Eclipse dataset, a simpler approach was followed:
As the times at which applications are executed relative to each other varies between the training, test, and validation set, the augmentation strategy for this dataset consists of shifting the data channels in time by a randomized amount.

\subsection{Thresholding and Anomaly Scoring}
Thresholding for all transformer-based methods is performed using the well-known Streaming Peak-Over-Threshold (SPOT) method~\cite{siffer2017anomaly}.
This method constitutes an Extreme-Value-Theory-based approach to finding a suitable threshold using an estimate of the risk of outliers $q$ and a set of known anomaly-free data.
In this work, the first 10\% of the training datasets are supplied to SPOT for the purpose of threshold determination.

\subsection{Combined Detection}
In the combined detection setting, the predictions of the DBSCAN-based detection and the transformer-based detection are overlaid using a logical-or operation to form a combined anomaly prediction.
As the predictions made by the T-DBSCAN approach are binary, dependent on cluster membership of the individual data channels, the results are combined after thresholding is applied to the transformer predictions.
Figure~\ref{fig:offline-detection-pipeline} shows the flow of data in the DeepHYDRA approach.\newline
As is visible, the data paths for the DBSCAN-based detection and the transformer-based detection do not have dependencies between each other.
This means that the DeepHYDRA approach can benefit from available parallelism in both a combined GPU/CPU computation, where the transformer-based model would run on the GPU while DBSCAN is run on the CPU, as well as CPU-only parallelism, as is the case for the projected deployment scenario at ATLAS.\newline
The T-DBSCAN-based component of DeepHYDRA, and as a consequence the combined detection, does not require training additional to training the long-term detection model, only the selection of its parameters $\epsilon$ and min\_samples.

\section{Experimental Methodology}
\label{sec:experiments}
In the following, we lay out the datasets used, and detail the method of creation of the novel datasets, as well as the labelled SMD machine-1-1 training dataset.
Following that is a discussion of the evaluation metrics used to gauge the performance of the stand-alone anomaly detection and the combined anomaly detection methods.
This is lastly followed by a detailed description of the parameters used in the analysed models and the applied experimental procedure.
\subsection{Baseline Algorithms}
We include a selection of non-ML-based algorithms to compare the analysed ML-based algorithms against.
These methods were recently outlined as suitable baselines for comparing the performance of time-series anomaly detection algorithms against~\cite{wu2021current}.
These algorithms are detailed in the following.\newline
Wu and Keogh recently showed that many contemporary deep-learning-based time-series anomaly detection methods are outperformed using very simple  algorithms, which they name one-liner algorithms~\cite{wu2021current}.
They present six such methods, of which the methods 3 and 4, which are both based on absolute differences, are used as baselines in this work.
The reasoning behind this is that the HLT datasets contain anomalies that result in both negative and positive discrete differences, making these two methods the most suitable.\newline
The third baseline algorithm used is a version of the MERLIN algorithm~\cite{nakamura2020merlin} implemented by Ferdinand Rewicki.~\cite{rewicki2022pymerlin}.
MERLIN is based on distances between data subsequences called discords, with large discords constituting potential anomalies~\cite{rewicki2023is}.
The MERLIN method is only suitable for univariate data, and is thus applied to each data channel individually.
\subsection{Datasets}
The deep-learning-based methods are evaluated on both the presented novel datasets, the Eclipse dataset.~\cite{aksar2023prodigy}, as well as the machine-1-1 subset of the SMD~\cite{su2019robust}.
As no labelled training dataset exists for the latter, one is created based on a small test dataset subsample containing anomalies.
The 2400 samples of this dataset are perturbed using APP~\cite{wen2021time}, as is done in the data augmentation method described in section~\ref{subsec:data-augmentation}.
This is an advantage for the SMSE-loss-based informer, as it is actually provided test data, even though the unlabelled training data is reduced by an amount equal to the additional data provided through the labelled training dataset.
As we use this dataset to judge the basic validity of the proposed semi-supervised training methodology, we deem this acceptable, as the tests on the newly-introduced datasets use an actually separate labelled training dataset that contains no test or validation data.
A possible increase in performance observed in the machine-1-1 dataset but not in the in-house dataset might hint at the performance increase only stemming from being exposed to actual test data, allowing us to discern whether this approach has merit or not.\newline
The novel test, validation, and labelled training datasets, which we will refer to as HLT datasets in the remainder of this work, are created by taking known anomaly-free data and injecting it with six types based on the well-defined anomaly type classification described by Lai et al.~\cite{lai2021revisiting}.
Of the described anomaly categories, global and contextual point outliers, collective global outliers, and collective trend outliers are included, and are introduced into the data based on code provided by Lai et al. as part of the TODS framework.
In a hierarchically-connected computing cluster scenario, global and contextual point outliers and collective outliers model more critical failure modes such as problems in the underlying inter-rack interconnect affecting whole server racks at once.\newline
In addition, two new categories named persistent global and persistent contextual outliers are introduced, which follow the same contextuality principles as their point-wise counterparts, but evolve over a stretch of time.
This sets them apart from collective global outliers, in which the data is completely replaced by synthetic data not present in the model, and collective trend outliers, which follow a discernible trend throughout the anomaly.
The amount of perturbation applied to the data is based on a Brownian bridge, with data becoming slowly more anomalous, and fading into normal data at the end of the anomaly window.
As a result, anomalies of this type have neither a sharp transitions between normal and anomalous data nor do they represent consistent, easily learnable patterns.
The gradually evolving persistent global and persistent contextual anomalies, along with the collective trend anomalies described by Lai et al. model a range of slowly evolving failures, such as slowdowns in data collection and processing upstream form the system, or cooling problems causing whole server racks to thermal throttle.
All anomalies are labelled based on their type in order to allow for analysis of the model performance on a per-category basis.\newline
For training and testing of the learning-based models, the anomaly-injected data are reduced to the per-subgroup median and standard deviation, forming the reduced HLT datasets.
This reduces the dimensionality of the data from around 3700 to 102, stemming from 51 active server racks.
There are never more than 1901 of these data channels active at a time, resulting in a sparsity of around 49 \%.
These datasets are min-max normalised to the range $[0, 1]$, with scaling based on the training dataset.\newline
A different dataset is created to evaluate the combined T-DBSCAN and learning-based anomaly detection methods.
In addition to the anomalies mentioned above, this dataset contains per-subgroup anomalies affecting only few channels or only a single channel.
These anomalies are specifically crafted to not cause a significant change in the median and standard deviation of the per-subgroup reduced data in order to render their detection in the reduced data difficult and specifically test the capabilities of the T-DBSCAN anomaly detection algorithm.
In a computing cluster, such anomalies that only affect a very small number of channels or a single channel model failure modes such as faulty or badly seated cables, power supply failures, or hardware/software faults in individual processing nodes.
This dataset, named the unreduced HLT dataset, is not normalised.
This way, the combined detection strategy can be evaluated in a simulated streaming scenario, where the data processed by the deep-learning-based models is reduced and normalised on-the-fly.
The Eclipse datasets presented by Aksar et al.~\cite{aksar2023prodigy} consists of a set of telemetry records obtained from the Eclipse computing cluster located at the Sandia National Laboratories.
It cluster consists of 52 compute nodes connected by 13 switches, with each switch being connected to four nodes.
The telemetry data that makes up the dataset is provided by the Lightweight Distributed Metric Service (LDMS)~\cite{aaziz2021integrated}.
These channels together give an overview over the memory allocations, information about processes, interrupts, and other metrics that might be important to assess if a monitored program is in an anomalous state.
The datasets provided by the authors of the original work contains monitoring data from the applications LAMMPS, sw4, sr4Lite, and ExaMiniMD as representatives of common HPC workloads~\cite{aksar2023prodigy}.
The anomalies present in the provided test dataset are synthetic anomalies, but of a different nature than those present in the HLT datasets:
While in our work, anomalies are introduced post-hoc into historic monitoring data, Aksar et al. use the HPAS library~\cite{ates2019hpas} to simulate anomalies during the runtime of the monitored applications~\cite{aksar2023prodigy}.\newline
As the DeepHYDRA approach is tailored towards online monitoring of many nodes running in parallel, a reshaping of the dataset is required.
The individual records contain many overlapping timestamp ranges and can thus be aligned with each other to create much higher-dimensional and sparser datasets with less individual samples.
The resulting dataset is subgroup-reducible, making it suited for the application of the DeepHYDRA method, where the subgroups are not racks as in the HLT dataset but the individual metrics reported for each application.\newline
To reduce the otherwise extreme dimensionality, components that have contributions close to zero over all of the datasets are identified via PCA~\cite{jolliffe2002springer} and removed before the alignment process.
Labelled training and validation sets are created by randomly combining records from the original training and test datasets.
The end result of this reshaping are datasets with a dimensionality of 12800 compared to the original 159 but of a much lower combined sample count. 
The per-subgroup reduced datasets created from this to train the long-term detection models have a dimensionality of 640.\newline
It is of note here that this process results in an overlap of anomalous periods, thus increasing the number of concurrent anomalies occurring in a given anomalous period, meaning that the detection task is potentially simpler compared to the original datasets.
This should be taken into consideration when comparing the results of our evaluations with the results given for Prodigy model.
The reshaped Eclipse dataset nonetheless serves as a testing ground to assess the generalisability of the DeepHYDRA approach to other datasets where the opportunity of per-subgroup reduction exists.
\subsection{Detection Performance Metrics}
The reported metrics include the area under receiver operating characteristic curve (AUC-ROC)~\cite{powers2008evaluation}, F1 score, precision, and recall~\cite{goutte2005probabilistic}, as commonly used in the literature.
The receiver operating characteristic curve is a curve in two-dimensional space with the false-positive rate and the true-positive rate of a given binary classifier as its axes~\cite{powers2008evaluation}.
This metric effectively gives insights into how the choice of threshold affects possible false-positive and true-positive rates a classifier can achieve, as one generally has to trade one for the other.
The area under the receiver operating characteristic, the AUC-ROC, is then defined as area under this curve, essentially condensing the performance of a classifier over all possible thresholds into a score.
Precision is defined as the ratio of true positive to all predicted-positive predictions, while recall is the ratio of predicted-positive to actual-positive samples~\cite{goutte2005probabilistic}.
The F1 score, defined as the geometric mean between precision and recall, is useful as it combines the two into a single metric, equally weighing precision and recall, making it easier to ensure that thresholds are chosen in a manner that delivers both high precision and recall without sacrificing one for the other.
These metrics are reported based on the binarised adjusted instance corrected predictions~\cite{doshi2022tisat}, also as is common in the literature, even though it is known that reporting the AUC-ROC for binarised predictions can potentially lead to misleading results~\cite{muschelli2020roc}.
To address the shortfalls of these metrics, we additionally provide the Matthews Correlation Coefficient (MCC), which both is an adequate metric for binarised predictions and, unlike the F1 score, is resilient to class imbalance~\cite{chicco2020advantages,chicco2023matthews}.
As both the HLT datasets and the SMD datasets contain more than one anomaly, the UCR anomaly score, which is based on the assumption of a single anomaly per dataset, is not applicable~\cite{rewicki2023is}.\newline
For the univariate baseline algorithms, the per-channel predictions are reduced to predictions for the dataset as a whole using a logical-or operation over all channels.
The high-dimensional HLT datasets have a large overall variance in terms of difficulty of the detection of the included anomalies.
This is addressed in the reported results for the baseline algorithms by filtering the channel predictions by individual MCC and then choosing the threshold for which the overall produced scores are the highest.\newline
For the modified Eclipse dataset~\cite{aksar2023dataset}, we elect to omit the application of the adjusted instance-based processing of predictions, as this dataset contains only two anomaly windows, meaning that this process would potentially introduce an extreme skew in the reported metrics.

\subsection{Computational Intensity and Memory Footprint Study}
The computational intensity and memory requirements of the analysed models are retrieved or measured using a set of different approaches, depending on what is most suitable.
For the machine-learning-based detectors, we make use of the activation count and FLOP retrieval methods offered by the fvcore Python module published by Facebook AI Research (FAIR)~\cite{facebook2023fvcore}.
Where necessary, custom operator handlers were added to cover the operators not handled by this module by default, such as the GRU layers of OmniAnomaly~\cite{su2019robust}\footnote{The implementation of the custom handlers are made available under \url{https://github.com/UniHD-CEG/torchprofilingutils}}.
Parameter counts are obtained using the torchinfo module by Tyler Yep~\cite{yep2020torchinfo}.\newline
The computational intensity of the non-machine-learning-based methods, especially DBSCAN and MERLIN, is highly data dependent.
To obtain reliable computational intensity data for the traditional approaches, we employ the the pylikwid module~\cite{rrzehpc2021pylikwid}, a wrapper around the popular performance monitoring suite likwid~\cite{treibig2010likwid}.
To obtain the memory footprint the scikit-learn-based code, namely the reduction and DBSCAN, we make use of the Python module SkLearn2PMML developed by Villu Ruusmann, which offers memory footprint retrieval for scikit-learn objects~\cite{jpmml2024sklearn2pmml}.
For the one-liner methods, the memory footprint is calculated based on the number of parameters per channel, which is 1 for Method 3 and 3 for Method 4.
For MERLIN, the sizes of the arrays allocated in the functions executed by the py-merlin implementation during detection~\cite{rewicki2022pymerlin} are registered.
We report both a size for sequential application of MERLIN, the maximum of the sizes for each channel, and the sum of the sizes of allocated arrays, which represents the memory footprint of parallel application of MERLIN.
\subsection{Model Parameters}
The parameters for T-DBSCAN were experimentally determined based on the training data.
For OmniAnomaly, DAGMM, USAD, and TranAD, the parameters established by Tuli et al.~\cite{DBLP:journals/corr/abs-2201-07284} for training on SMD were adapted to accommodate the greater dimensionality of the HLT datasets.
The parameters used in the Informer-MSE and Informer-SMSE models were determined empirically with the exception of the Informer-MSE model for the HLT datasets, where better suited parameters were found using a hyperparameter search using the Asynchronous Successive Halving Algorithm~\cite{li2020massively} provided by the RayTune library~\cite{liaw2018tune}.\newline
The exact hyperparameters used for each model-dataset combination are available in our model repository.
\subsection{Experimental Procedure}
All investigated models are tested in a randomised seed trial using 10 different seeds, with the reported metrics corresponding to the best-performing seed.\newline
The first round of testing is performed on the machine-1-1 dataset, which serves to validate the existing methods as well as to evaluate the impact of semi-supervised training on a known dataset.
This is then followed by testing on the HLT dataset, where additionally the impact of data augmentation is investigated.
To find the best combination of data augmentation for each model, a total of 150 augmentation settings are tested, varying in the amount of scaling applied, the magnitude of APP parameters, and the proportion of augmented training data to unaugmented training data.
This is done using a fixed random seed to ensure comparability between the different augmentation parameter combinations, with a random seed trial performed on the best candidate augmentation.\newline
Lastly, the best-performing candidates for each model based on the previously conducted augmentation study are evaluated on the unreduced HLT dataset including intra-subgroup outliers.
For this set of experiments, the machine learning-based methods are only tested using the aforementioned on-the-fly  reduction and normalization, as training the models on the full unreduced dataset without any reduction is not feasible on available hardware.
In order to gauge the impact each of the constituent methods of DeepHYDRA have on the detection performance, the candidate transformer methods and T-DBSCAN are both tested as stand-alone algorithms as well as in the proposed hybrid detection setting.
This approach is also followed for the Eclipse dataset.
For this dataset, all models are trained on eight versions of the modified Eclipse dataset to which the described randomized time-shift-based augmentation strategy was applied.
\newline
The modified Eclipse dataset~\cite{aksar2023dataset} serves as a test of the generalisability of the DeepHYDRA approach.
We choose to not perform a hyperparameter search and only apply the simple time-shift-based augmentation strategy, performing no search for optimal augmentation parameters.
The rationale behind this evaluation approach is to gauge how easily DeepHYDRA-based detection can be applied to previously unseen datasets. \newline
The parameters of the included one-liner algorithms are determined per channel to produce the best possible overall performance.
The configuration of the training and benchmark system used in this work are made available in our model repository.

\section{Results}
\begin{table*}
	\centering
	\subfloat[][SMD Machine-1-1 Test Set\vspace*{0pt}]{
	\begin{tabular}{l|rrrrrrrr}
		\toprule
		Model  & \makecell{MCC  \\ (mean/std)} & \makecell{MCC \\ (best)} & \makecell{F1  \\ (best)} & \makecell{AUC-ROC  \\ (best)} & \makecell{Prec  \\ (best)} & \makecell{Rec  \\ (best)} & \makecell{FLOPs\\ per Sample} & \makecell{Parameters \&\\Activations} \\
		\midrule
		1LM 3 & - &   98 \% &  98 \% & \textbf{100 \%} & 97 \% &  \textbf{100 \%} & 38 & 38 \\
		1LM 4 & - & 99 \% &  99 \% & \textbf{100 \%} &  98 \% &   \textbf{100 \%} & 534 & 114 \\
		MRLN-S & - &  98 \% & 98 \% &  \textbf{100 \%} &  97 \% &  \textbf{100 \%} &  $\num{5.77e8}$&  $\num{5.6e6}$ \\
		MRLN-P & - &  98 \%&  98 \% &  \textbf{100 \%} &  97 \% &  \textbf{100 \%} &  $\num{5.77e8}$ &  $\num{2.1e8}$ \\
		OA &  98.2 \% $\pm$ 0 \% &  98 \% &   98 \% &  86 \% & 97 \% & \textbf{100 \%} & $\num{1.82e4}$ & $\num{1.97e4}$ \\
		DAGMM & 96 \% $\pm$ 0.8 \% &  97 \% &98 \% & 94 \% &   96 \% & \textbf{100 \%}  & $\num{1.02e4}$ & $\num{1.1e4}$ \\
		USAD &  93.2 \% $\pm$  2.2 \% & 98 \% &  98 \% & 93 \% & 96 \% &  \textbf{100 \%} & $\num{1.69e4}$ & $\num{1.13e4}$ \\
		TranAD &  90.7 \% $\pm$  1.9 \% &   95 \% &  95 \% &   93 \% & 91  \% & \textbf{100 \%}  & $\num{1.02e6}$ & $\num{1.73e5}$ \\
		IF-MSE&  97.8 \% $\pm$  2.3 \% & \textbf{100 \%} & \textbf{100 \%}  & \textbf{100 \%}  & \textbf{100 \%} & \textbf{100 \%} & $\num{7.06e8}$ & $\num{2.08e7}$ \\
		IF-SMSE &  99 \% $\pm$  0.5\% &   \textbf{100 \%} & \textbf{100 \%} &  \textbf{100 \%} & 99 \% & \textbf{100 \%}  & $\num{7.06e8}$ & $\num{2.08e7}$ \\
		\bottomrule
	\end{tabular}
	\label{table:results-transformer-evaluation-smd}
	\vspace*{2pt}
	}\\
	\vspace*{2pt}
	\subfloat[][Pre-Reduced HLT Test Set\vspace*{0pt}]{
	\setlength{\tabcolsep}{4.5pt}
	\centering
	\begin{tabular}{l|rrrrrrrr}
	\toprule
		Model  & \makecell{MCC  \\ (mean/std)} & \makecell{MCC \\ (best)} & \makecell{F1  \\ (best)} & \makecell{AUC-ROC  \\ (best)} & \makecell{Prec  \\ (best)} & \makecell{Rec  \\ (best)} & \makecell{FLOPs\\ per Sample} & \makecell{Parameters \&\\Activations} \\
		\midrule
		 1LM 3 & -  & 86 \% & 88 \% & 94 \% & 85 \% & 91 \% & 102 & 102 \\
		 1LM 4 & -  & 85 \% & 87 \% & 94 \% & 83 \% & 91 \%  & 1142 & 306 \\
		 MRLN-S & -  & 78 \% & 81 \% & 90 \% & 79 \% & 83 \%  & $\num{1.4e8}$ &$\num{5.4e6}$ \\
		 MRLN-P & -  & 78 \% & 81 \% & 90 \% & 79 \% & 83 \%  & $\num{1.4e8}$ & $\num{5.51e8}$ \\
		OA &  65.4 \% $\pm$ 1.2 \% &  68 \% &  73 \% &  84 \% &  74 \% &  73 \%  & $\num{2.64e4}$ & $\num{2.8e4}$ \\
		OA-AGM & 76 \% $\pm$  4 \%  &  82 \% &  84 \% &  94 \% &  78 \% &  93 \%  & $\num{2.64e4}$ & $\num{2.8e4}$ \\
		DAGMM &  91.3 \% $\pm$ 1.8 \% &  93 \% &  94 \% &  96 \% &  95 \% &  94 \%  & $\num{2.59e4}$ & $\num{2.76e4}$ \\
		DAGMM-AGM & 90.9 \% $\pm$  2.1 \%  &  \textbf{94 \%} &  \textbf{95 \%} &  96 \% &  96 \% &  94 \%  & $\num{2.59e4}$ & $\num{2.76e4}$ \\
		USAD &   88.2 \% $\pm$  5.8 \%  & 92 \% &  93 \% &  96 \% &  97 \% & 94\% & $\num{4.25e4}$ & $\num{2.83e4}$ \\
		USAD-AGM  &78.3 \% $\pm$  15.5 \%  &  89 \% &  91 \% &  93 \% &  \textbf{100 \%} &  86 \%  & $\num{4.25e4}$ & $\num{2.83e4}$ \\
		TranAD & 80.1 \% $\pm$  2.4 \% & 84 \% & 86 \% & 89 \% & 94 \% & 79 \%  & $\num{6.2e6}$ & $\num{9.97e5}$ \\
		TranAD-AGM & 90.3 \% $\pm$  2.4 \%  & 92 \% & 93 \% & 96 \% & 93 \% & 94 \%  & $\num{6.2e6}$ & $\num{9.97e5}$ \\
		IF-MSE & 40.2 \% $\pm$  35.1 \%  & 91 \% & 92 \% & 94 \% & 97 \% & 88 \%  & $\num{3.18e8}$ & $\num{2.99e7}$ \\
		IF-MSE-AGM  & 86.9 \% $\pm$  13.1 \%  & 93 \% & 94 \% & 97 \% & 94 \% & 95 \%  & $\num{3.18e8}$ & $\num{2.99e7}$ \\
		IF-SMSE & 92 \% $\pm$  0.6 \% & 93 \% & 94 \% & \textbf{97 \%} & 92 \% & 95 \%  & $\num{7.16e8}$  & $\num{7.32e8}$ \\
		IF-SMSE-AGM & 90.1 \% $\pm$  0.8 \%  & 92 \% & 93 \% & \textbf{97 \%} & 90 \% & \textbf{96 \%}  & $\num{7.16e8}$ &  $\num{7.32e8}$\\
		\bottomrule
		\end{tabular}
	\label{table:results-transformer-evaluation-l1rate}
	\vspace*{2pt}
	}
	\caption{Comparison of mean MCC and best MCC, AUC-ROC, F1 score, precision, and recall for baseline methods and deep-learning-based methods on the SMD machine-1-1  dataset and the pre-reduced HLT dataset. Also shown are the FLOPs per sample required in inference as well as the required parameters and activations.
		Note that for the baseline methods and T-DBSCAN no mean MCC is given, as they are deterministic methods.
		While on the tested subset of the SMD dataset all tested methods show good performance, the non-learning-based methods struggle on the more complex HLT datasets, especially without pre-reduction. The introduced augmentation methodology noticeably improves the performance of most analysed models.}
	\label{table:results-all-0}
	\end{table*}
	
	\begin{table*}
	\ContinuedFloat
	\subfloat[][Unreduced HLT Test Set\vspace*{0pt}]{
	\setlength{\tabcolsep}{4.5pt}
	\centering
	\begin{tabular}{l|rrrrrrrr}
		\toprule
		Model  & \makecell{MCC  \\ (mean/std)} & \makecell{MCC \\ (best)} & \makecell{F1  \\ (best)} & \makecell{AUC-ROC  \\ (best)} & \makecell{Prec  \\ (best)} & \makecell{Rec  \\ (best)} & \makecell{FLOPs\\ per Sample} & \makecell{Parameters \&\\Activations} \\
		\midrule
		 1LM 3 & - &  82 \% & 84 \% &  92 \% & 82 \% & 86 \% & $\num{3.69e3}$ &  $\num{3.69e3}$ \\
		 1LM 4 & - &  84 \% & 86 \% & 93 \% & 84 \% & 87 \% & $\num{2.46e4}$ &  $\num{1.1e4}$ \\
		 MRLN-S & - &  72 \% & 75 \% & 86 \% & 76 \% & 75 \% & $\num{5.04e9}$ & $\num{5.4e6}$ \\
		 MRLN-P & - &  72 \% & 75 \% & 86 \% & 76 \% & 75 \% & $\num{5.04e9}$ &  $\num{1.99e10}$ \\
		T-DBSCAN & - &  79 \% & 80 \% & 83 \% & \textbf{99 \%} & 67 \% & $\num{5.25e7}$ &  $\num{1.33e4}$ \\
		OA &   70.7 \% $\pm$  3.9 \% &  77 \% & 79 \% & 91 \% & 85 \% & 85 \% & $\num{6.08e7}$ & $\num{2.8e4}$ \\
		DH-OA &  78.7 \% $\pm$  2.6 \%  & 82 \% & 84\% & 95 \% &87 \% & 94 \% & $\num{1.33e8}$ &$\num{4.14e4}$ \\
		DAGMM & 78.8 \% $\pm$ 2.1 \% & 80 \% & 81 \% & 85 \% & 98 \% & 71 \% & $\num{6.08e7}$ &$\num{2.76e4}$ \\
		DH-DAGMM &   89.7 \% $\pm$ 2.6 \% & \textbf{92 \%} & 92 \% & 94 \% & 98 \% & 89 \%  & $\num{1.33e8}$ & $\num{4.1e4}$ \\
		USAD & 75.3 \% $\pm$ 9 \% & 84 \% & 86 \% & 90 \% & \textbf{99 \%} & 85 \% & $\num{6.09e7}$ & $\num{2.83e4}$ \\
		DH-USAD  & 84.9 \% $\pm$ 5.2 \%  & \textbf{92 \%} & \textbf{93 \%} & 96 \% & \textbf{99 \%} & 95 \% & $\num{1.34e8}$ & $\num{4.16e4}$ \\
		TranAD & 78.9 \% $\pm$  0.9 \% & 80 \% & 83 \% & 89 \% & 85 \% & 81 \% & $\num{6.7e7}$ &  $\num{9.97e5}$ \\
		DH-TranAD & 86.5 \% $\pm$  1 \%  & 88 \% &  89\% & 95 \% & 86 \% & 93 \% & $\num{1.2e8}$ &  $\num{1.01e6}$ \\
		IF-MSE &81.8\% $\pm$  1.1 \% & 84 \% & 86 \% & 92 \% & 85 \% & 86 \% & $\num{3.79e8}$ &  $\num{2.99e7}$ \\
		DH-MSE & 88.1 \% $\pm$  0.9 \% & 89 \% &  91 \% & \textbf{97 \%} & 86 \% & \textbf{96 \%} & $\num{4.3e8}$ & $\num{2.99e7}$ \\
		IF-SMSE & 83.7 \% $\pm$  0.5 \% & 85 \% & 86 \% &  93 \% & 88 \% & 87 \% & $\num{7.77e8}$ &  $\num{7.32e8}$ \\
		DH-SMSE & 90.2 \% $\pm$  0.7 \%  & 91 \% & 92 \% & \textbf{97 \%} & 90 \% & \textbf{96 \%} & $\num{8.3e8}$ &  $\num{7.32e8}$ \\
		\bottomrule
	\end{tabular}
	\label{table:results-transformer-evaluation-combined-hlt}
	\vspace*{2pt}
	}
	\vspace*{2pt}
	\subfloat[][Modified Eclipse Test Set\vspace*{0pt}]{
		\setlength{\tabcolsep}{4.5pt}
		\centering
		\begin{tabular}{l|rrrrrrrr}
			\toprule
			Model  & \makecell{MCC  \\ (mean/std)} & \makecell{MCC \\ (best)} & \makecell{F1  \\ (best)} & \makecell{AUC-ROC  \\ (best)} & \makecell{Prec  \\ (best)} & \makecell{Rec  \\ (best)} & \makecell{FLOPs\\ per Sample} & \makecell{Parameters \&\\Activations} \\
			\midrule
			1LM 3& - &   \textbf{100  \%} & \textbf{100 \%} & \textbf{100  \%} &  \textbf{100  \%} &  \textbf{100  \%} & $\num{1.28e4}$ &  $\num{1.28e4}$ \\
			 1LM 4 & - &   \textbf{100  \%} &  \textbf{100  \%} &  \textbf{100  \%} &  \textbf{100  \%} & \textbf{100  \%} & $\num{1.55e5}$ & $\num{3.84e4}$ \\
			 MRLN-S & - &  88 \% & 95 \% & 93 \% & 90 \% & \textbf{100 \%} & $\num{5.1e11}$ & $\num{3.23e6}$ \\
			 MRLN-P & - & 88 \% & 95 \% & 93 \% & 90 \% & \textbf{100 \%} & $\num{5.1e11}$ &  $\num{2.07e9}$ \\
			T-DBSCAN & - & 95 \% & 97 \% &  98 \% & \textbf{100 \%} & 95 \% & $\num{1.24e7}$ &  $\num{1.17e5}$ \\
			DH-DAGMM &   91.3 \% $\pm$  0.2 \% & 92 \% & 96 \% & 96 \% & 95 \% & 97 \% & $\num{1.26e7}$ & $\num{2.85e5}$ \\
			DH-USAD & 91.0 \% $\pm$ 0.1 \% & 91 \% & 96 \% & 96 \% & 96 \% & 96 \% &$\num{1.27e7}$ &$\num{2.88e5}$ \\
			DH-TranAD & 91.3 \% $\pm$ 0.0 \% & 91 \% & 96 \% & 96 \% & 95 \% & 97 \% & $\num{2.85e10}$ &  $\num{8.19e7}$ \\
			\bottomrule
		\end{tabular}
		\label{table:results-transformer-evaluation-combined-eclipse}
		\vspace*{2pt}
	}
	\caption{Continued. Comparison of mean MCC and best MCC, AUC-ROC, F1 score, precision, and recall for baseline methods and deep-learning-based methods on the unreduced HLT dataset and the modified Eclipse dataset. Also shown are the FLOPs per sample required in inference as well as the required parameters and activations.
		Note that for the baseline methods and T-DBSCAN no mean MCC is given, as they are deterministic methods.
		While the transformer-based methods show consistently good performance on the pre-reduced and unreduced datasets, the less computationally intensive models DAGMM and USAD outperform them on the HLT dataset using DeepHYDRA.
		For Eclipse, while none of the tested method reach the baseline using one-liner methods, the T-DBSCAN method shows the best result for real-time detection.
		It is of note here that DeepHYDRA was not optimised for this detection task, and this evaluation merely serves as a test of the generalisibility of DeepHYDRA to other HPC detection tasks.}
	\label{table:results-all}
\end{table*}
\subsection{SMD Datasets}
Table~\ref{table:results-transformer-evaluation-smd} shows the results of our analyses on the machine-1-1 dataset.
All of the algorithms show perfect recall, resulting in the main differentiator in performance being the achieved precision.
Of the transformer-based algorithms, the Informer-models are closely tied in delivering the best performance, with Informer-MSE achieving a perfect score in all metrics.
Overall, the results for the investigated transformer-based models are comparable with the baseline methods, with the one-liner method 4 achieving an almost perfect score as well.
Through the analysis of a wide range of subsequence lengths, the MERLIN algorithm manages to capture both the short-term and long-term anomalies contained in the machine-1-1 test set, resulting in it outperforming all of the considered transformer-based methods.

\subsection{Pre-Reduced HLT Datasets: Augmentation and Semi-Supervised Learning Evaluation}
On the larger, higher-dimensional reduced HLT dataset shown in Table~\ref{table:results-transformer-evaluation-l1rate} however, the gap in performance increases noticeably.
While all of the baseline algorithms perform remarkably well, there is a large deviation between individual channels, which can be somewhat alleviated by filtering out channels with low detection performance.
The MERLIN algorithm still produces fairly good results, considering that the dataset contains many anomalies per channel, which should be a limiting factor for the particular implementation where only the highest discord per subsequence length is reported.
The best performing algorithms for this dataset are all transformer-based models.
The Informer-MSE and Informer-SMSE models are tied for the highest AUC-ROC, while DAGMM trained on augmented data leads in terms of MCC and F1-score.
There is an overall observable trend of the models trained on augmented data outperforming their unaugmented-data-trained counterparts in the combined metrics.
Additionally, both the introduced augmentation strategy and semi-supervised loss seem to make the informer-based detection much more stable in performance over the used random seeds.

\subsection{Unreduced HLT Datasets: DeepHYDRA Evaluation}
With augmentation and semi-supervised learning established as beneficial methods to boost detection performance, we turn our focus to the evaluation of performance of the proposed DeepHYDRA workflow.
Table~\ref{table:results-transformer-evaluation-combined-hlt} shows the stand-alone performance of the baseline models and the learning-based methods, as well as the performance of the combined DeepHYDRA strategy.
The results show clearly that the combination of T-DBSCAN-based and learning-based detection as proposed for DeepHYDRA achieves the desired outcome:
Across all learning-based models, the combined detection vastly increases performance, with an up to 12 \% increase in MCC visible for DAGMM.
While the baseline methods still show fairly good performance, the combined approach is able to greatly outperform them.

\subsection{Modified Eclipse Dataset}
The results for the modified Eclipse dataset, shown in Table~\ref*{table:results-transformer-evaluation-combined-eclipse}, are rather surprising, as the one-liner baseline methods show perfect detection across all metrics.
A possible explanation for this is that because of the high dimensionality of the dataset, finding optimal thresholds for the method for each of the individual data channels and combining the results enables perfect detection.
While the T-DBSCAN method on its own shows exceptionally good performance, the combination with the long-term anomaly detectors thus far fails to improve their performance aside from improving the recall.
\subsection{Computational Intensity and Memory Footprint Evaluation}
\begin{figure}
	\centering
	\includegraphics[alt={The scatterplots show MCC on the y-axes. The leftmost plots show inference FLOPs per sample on the x-axis, while the rightmost plots show the memory footprint on the x-axis. The x-axes all use logarithmic scale. Non-ML-based methods tend to be more computationally and space efficient, but have worse detection performance. MERLIN has a memory footprint comparable ML-based methods while showing lower detection performance. An overall trend is visible where DeepHYDRA is able to improve detection performance while only slightly increasing the computational cost and memory footprint of the stand-alone long-term detection methods.},width=0.8\columnwidth]{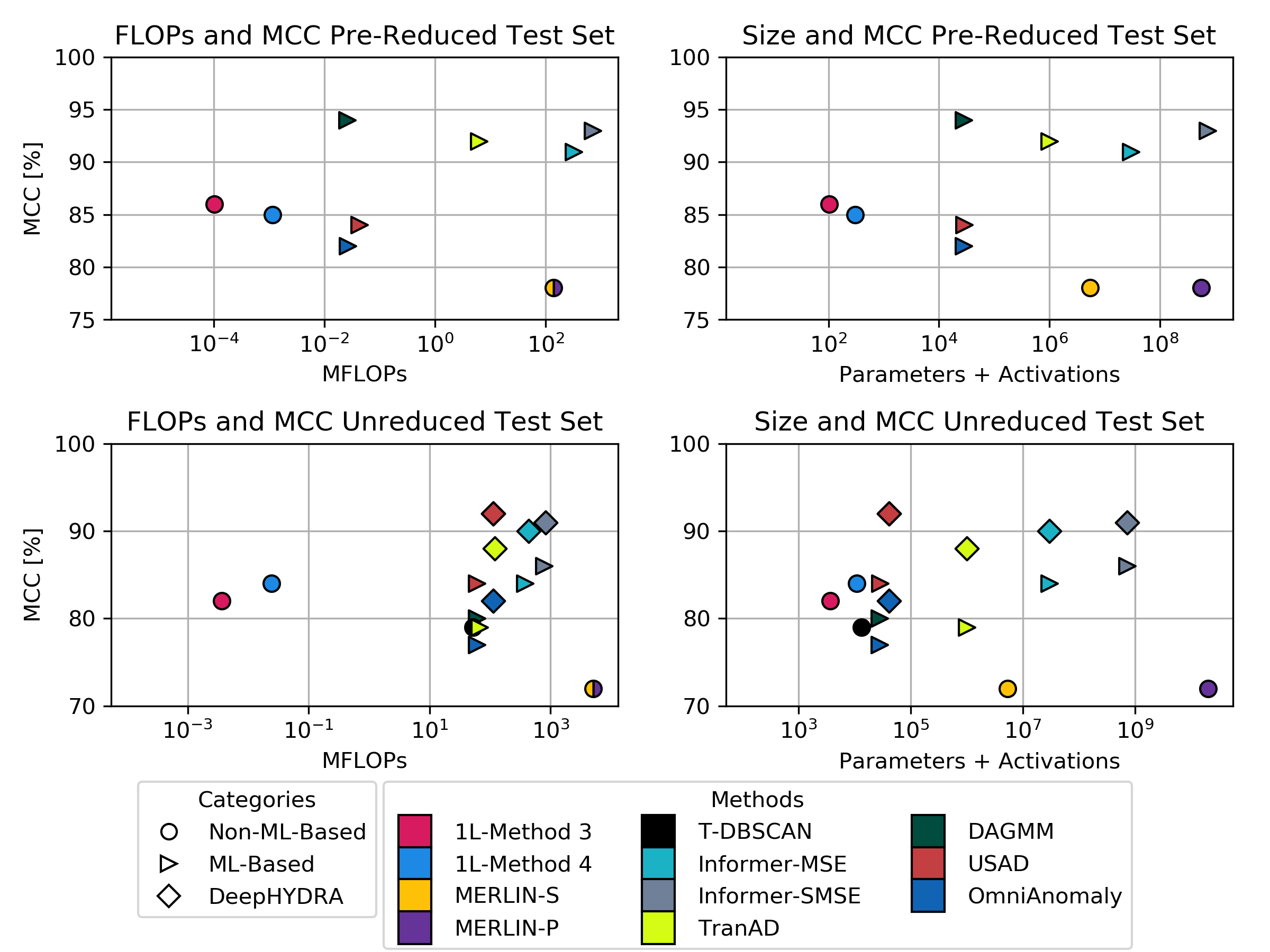}
	\caption{Inference FLOPs per sample and model size including activations over best MCC for pre-reduced and unreduced HLT datasets. The reported FLOPs for the machine-learning-based methods and hybrid methods include the FLOPs required in the reduction step.}
	\label{fig:efficiency-scatterplots}
\end{figure}
Figure~\ref{fig:efficiency-scatterplots} shows a comparison of the relationships between detection performance, computational intensity, and memory footprints, both on the pre-reduced and unreduced HLT datasets.
The three main results of our computational intensity and memory footprint investigation are readily apparent from these figures:
The first insight gained is that among the investigated methods, there is no clear correlation between computational intensity, memory footprint, and detection performance, with DAGMM and USAD consistently outperform the vastly larger and more computationally intensive transformer-based models. 
Secondly, the DeepHYDRA method is able to greatly boost detection performance of all tested learning-based approaches while only introducing a small increase in computational intensity and memory footprint.
A last and possibly surprising result is that the largest and most computationally intensive model is not one of the learning-based approaches, but MERLIN in the parallelized deployment scenario.
While it shows very good results on the machine-1-1 dataset, on the HLT sets, it is outperformed both in efficiency and overall performance by learning-based methods.
\subsection{Per-Category Performance}
\begin{figure}
	\centering
	\begin{subfigure}[b]{0.65\columnwidth}
		\centering
		\includegraphics[alt={This figure shows a single-timestep anomaly at the beginning of the period shown is only caught by the one-liner methods. A short drop to zero in one of the channels shortly after is detected by T-DBSCAN, and, as an extension, the DeepHYDRA methods, but none of the other models. A period of prolonged and more extreme anomaly at the end of the shown period is meanwhile identified by all of the shown methods},width=\textwidth]{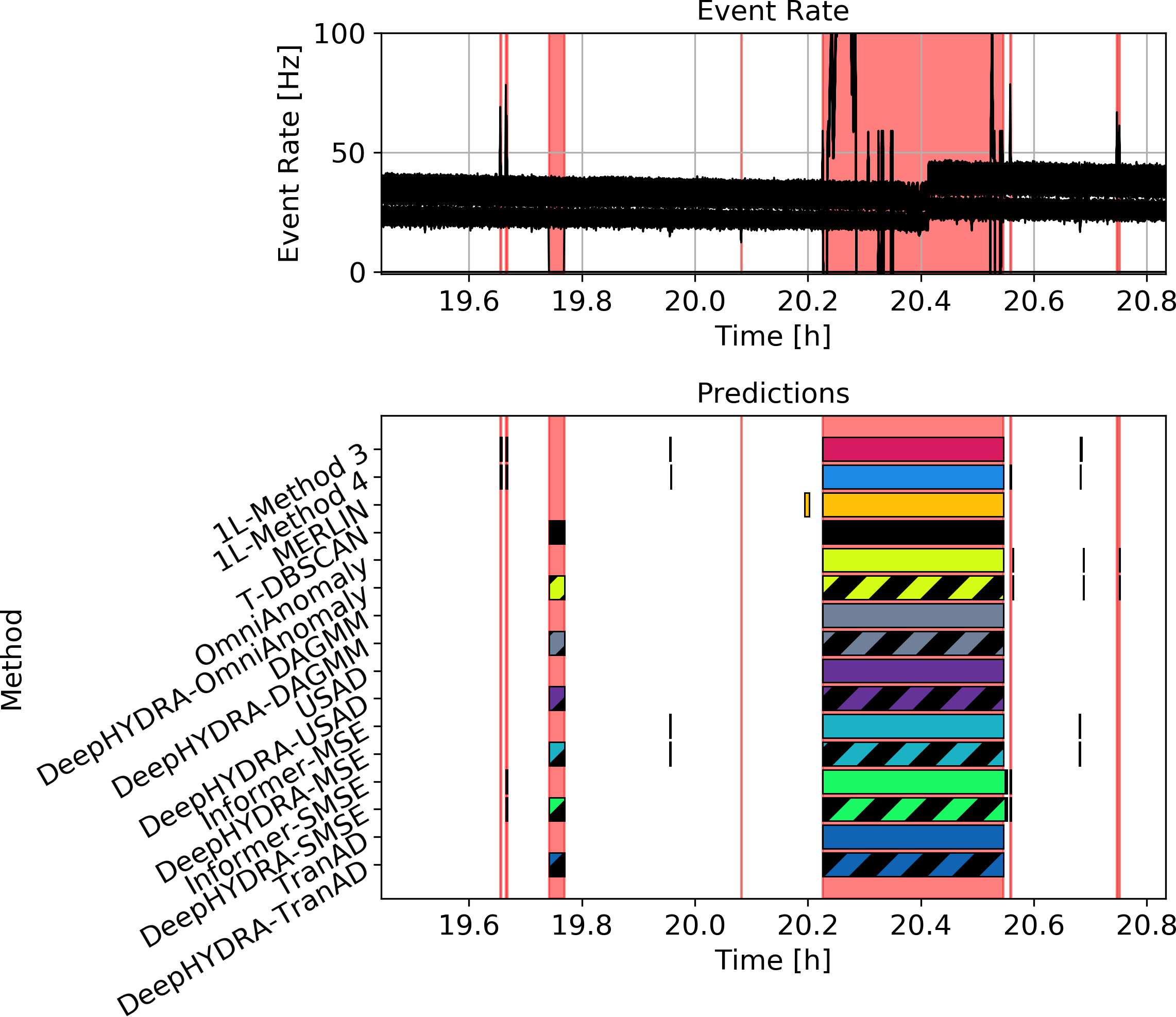}
		\caption{In this example, T-DBSCAN is able to identify a  short dropout of a node that eludes the other models, having only a miniscule effect on the reduced data stream. The long-term global anomaly following it meanwhile is caught by all investigated models.}
		\label{fig:prediction-comparison-long-term-difficult-1}
	    \vspace*{3pt}
	\end{subfigure}
	\begin{subfigure}[b]{0.65\columnwidth}
		\centering
		\includegraphics[alt={In this figure, a period of prolonged anomaly at the center of the plot that is not identified by T-DBSCAN but is identified by the one-liner methods and all long-term detection methods. OmniAnomaly and DeepHYDRA-OmniAnomaly meanwhile shows significant noise in its predictions that is absent from the other models. A subtle single-timestep anomaly following the period of prolonged anomaly is identified by none of the tested methods},width=\textwidth]{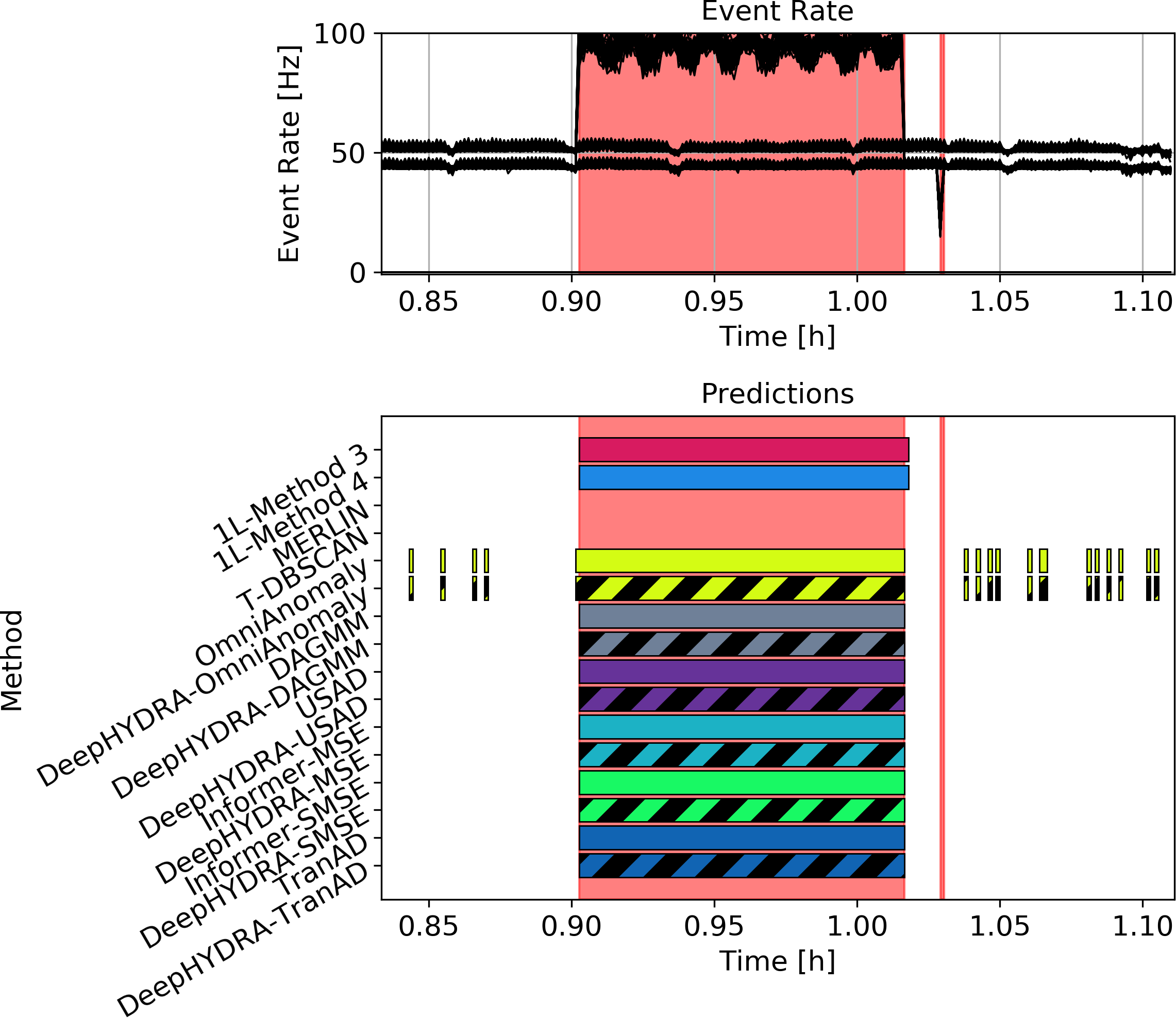}
		\caption{This example illustrates a shortfall of the T-DBSCAN method: During a period where all nodes in a subgroup show identical anomalies, it is unable to identify them as such. Here, the learning-based methods investigated for DeepHYDRA are able to identify the anomalous period.}
		\label{fig:prediction-comparison-long-term-difficult-0}
	\end{subfigure}
	\caption{Examples of the synergism between T-DBSCAN and the learning-based methods on the unreduced HLT dataset. Periods marked in red are anomalous. The bars in the lower subplots show predicted-positive periods for each method based on the used threshold.}
	\label{fig:prediction-comparison}
\end{figure}
\begin{figure}
	\centering
	\begin{subfigure}[b]{0.65\columnwidth}
		\centering
		\includegraphics[alt={A boxplot of the the MCC for the anomaly types of the unreduced HLT dataset for the evaluated models is shown. The medians and whiskers vary greatly between the models. While models such as DeepHYDRA-SMSE and DeepHYDRA-USAD show relatively consistent performance, others, such as T-DBSCAN, span almost the complete MCC range from 0 to 100 percent. The general observable trend is that Application of DeepHYDRA both improves median performance and performance consistency compared to the stand-alone long-term detection methods.},width=0.87\textwidth]{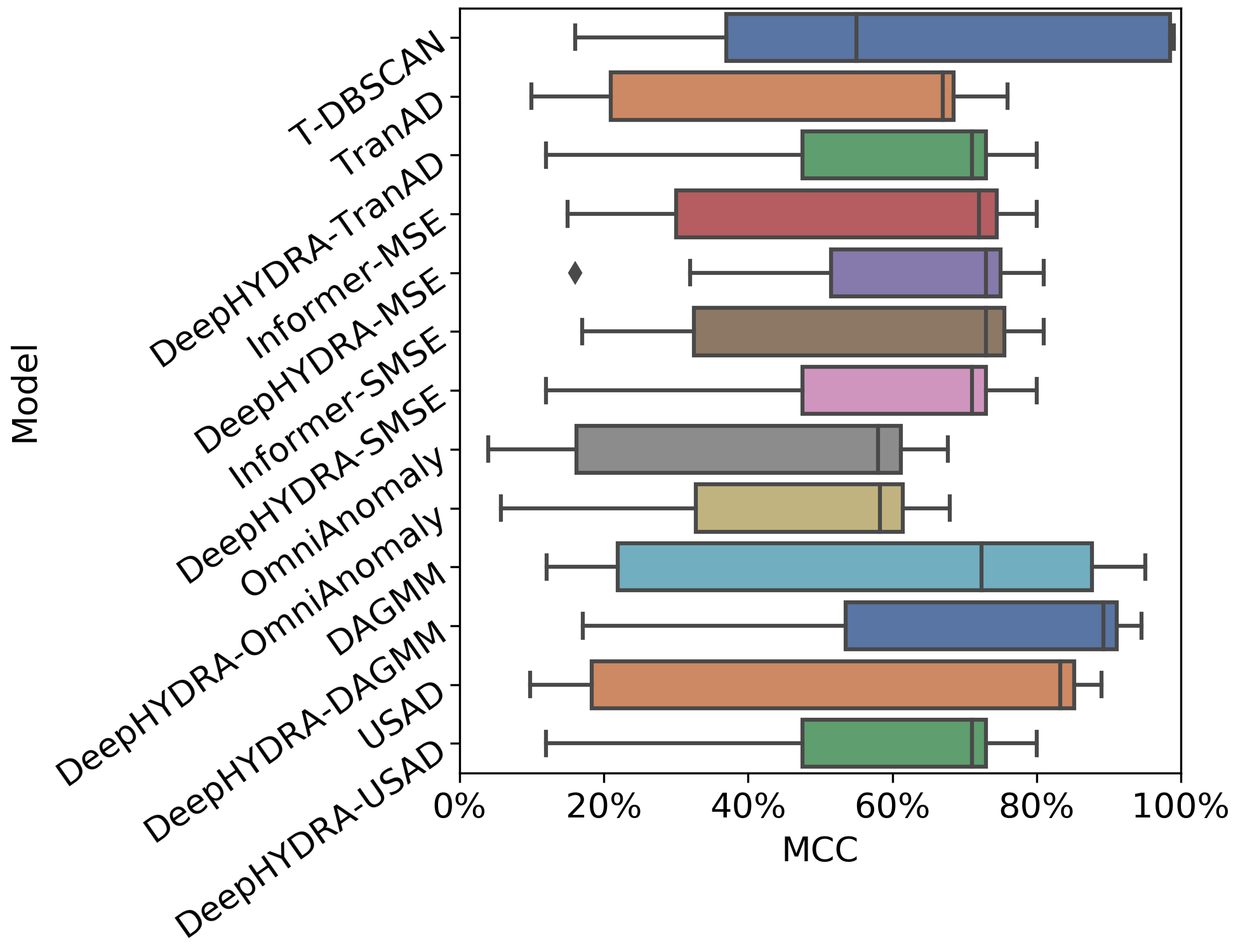}
		\caption{MCC by model.}
		\label{fig:combined-detection-boxplot-by-model}
	\end{subfigure}
	\begin{subfigure}[b]{0.65\columnwidth}
		\centering
		\includegraphics[alt={This figure shows the MCC variation for each of the anomaly types. Point Global and Contextual anomalies are shown to be consistently hard to detect, Persistent and Collective anomalies are comparatively easily identifiable. Intra-subgroup anomalies show the greatest variability in performance, spanning a large part of the MCC spectrum and showing low median MCC},width=0.87\textwidth]{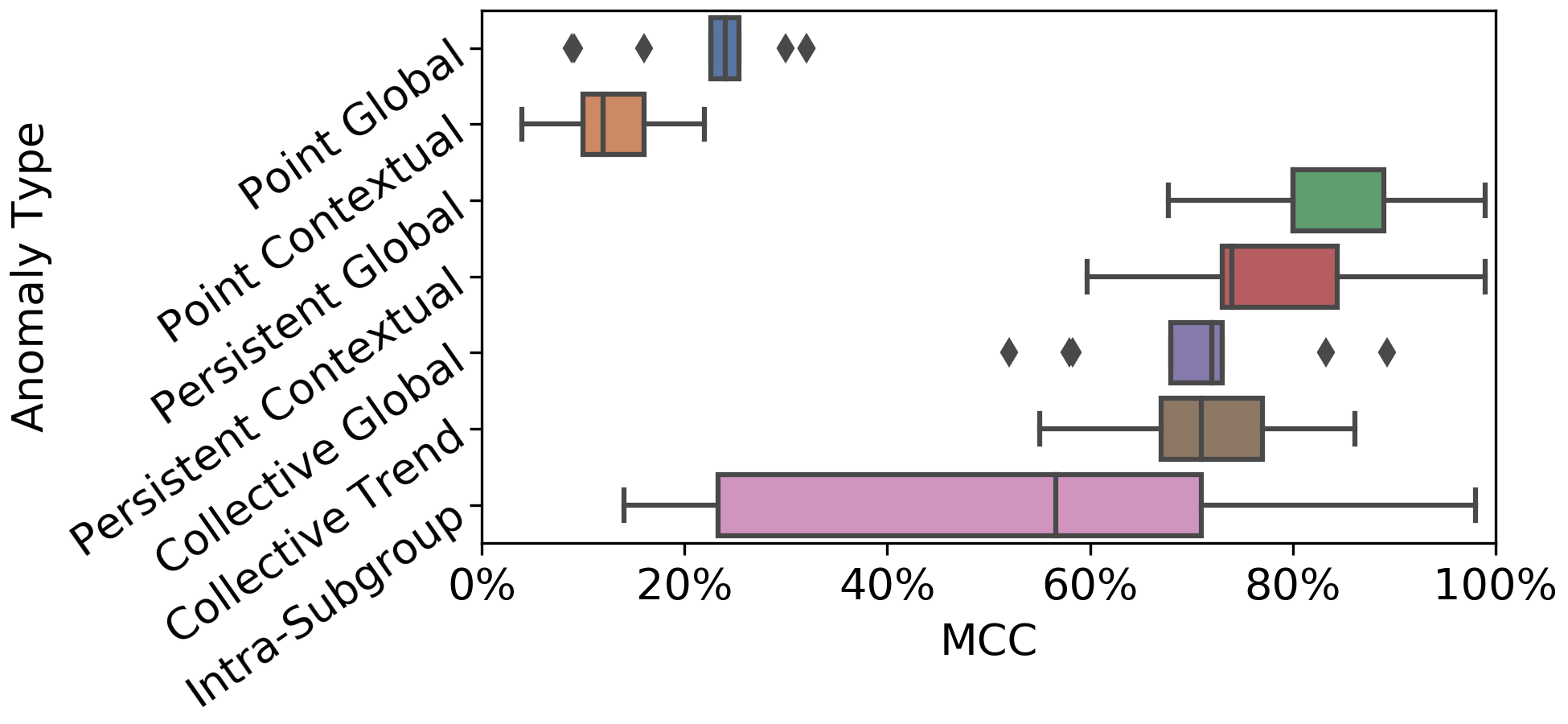}
		\caption{MCC by category.}
		\label{fig:combined-detection-boxplot-by-category}
	\end{subfigure}
	\caption{MCC for DeepHYDRA on the unreduced HLT dataset by model and category.}
	\label{fig:combined-detection-boxplots}
\end{figure}

Figure~\ref{fig:prediction-comparison-long-term-difficult-1} and~\ref{fig:prediction-comparison-long-term-difficult-0} shows two examples of different types of short-term and long-term anomalies.
These examples intend to show the specific strengths and weaknesses of the different parts of the proposed DeepHYDRA method:
While the short dropout of a single element at the beginning of the time period shown in Figure~\ref{fig:prediction-comparison-long-term-difficult-1} eludes the pure learning-based approaches, using DeepHYDRA, it is detected by T-DBSCAN.
Figure~\ref{fig:prediction-comparison-long-term-difficult-0} on the other hand serves as an example of a long-term global anomaly that impacts all of the channels in a subgroup at once, a behaviour which is impossible for T-DBSCAN to detect.
These types of anomalies are captured by the learning-based detection part of DeepHYDRA.\newline
The baseline methods can achieve remarkably good results through the previously mentioned filtering of channel predictions, outperforming the stand-alone transformer methods.
As these transformer-based methods process data that has been highly reduced, a process that unavoidably incurs loss of information, this comparatively lower performance is unsurprising.
In their DeepHYDRA counterparts however this performance loss is mostly mitigated by the T-DBSCAN method, resulting in these strategies outperforming all baseline methods.\newline
Figure~\ref{fig:combined-detection-boxplots} shows a comparison of the performance variation for the different anomaly categories for each of the investigated models, as well as the variation per category.
For this fine-grained evaluation we concentrate on the  MCC alone, with a detailed listing of all analysed metrics made available in our model repository.
It is of note that the median performances observed in these figures differ from the ones shown in Table~\ref{table:results-transformer-evaluation-combined-hlt}.
The reason for this is that the relative ratio of each anomaly type in the totality of samples marked as anomalous is not homogeneous.\newline
Observing Figure~\ref{fig:combined-detection-boxplot-by-model}, a clear take-away is that for most models, the combination with T-DBSCAN not only improves the mean performance, it also reduces the inter-class variation in performance. 
Comparing the results by anomaly category, it is evident that the performance for each kind of anomaly varies greatly.
The categories persistent contextual, persistent global, collective global, and collective trend are detected somewhat consistently by most of the models, while the categories Point Global and Point Contextual on the other hand prove challenging for all of the analysed methods.
The large span in performance for intra-subgroup anomalies is caused by most pure learning-based approaches performing very poorly on this category, while T-DBSCAN, and, as a result, the DeepHYDRA methods, capture them almost perfectly.

\section{Discussion}
Observing the reported results, especially the high scores of the simple one-liners, might lead one to conclude that the performance of the proposed deep learning and DBSCAN-based strategies is rather underwhelming, or that the datasets are simplistic.
However, based on the justified criticism of the evaluation methods used in many recent machine learning anomaly detection works laid out by Wu and Keogh~\cite{wu2021current}, considerable effort was put into optimizing the parameters of these methods to give a challenging baseline to compare our strategies to.
While the MERLIN algorithm is a well-established reliable anomaly detection algorithm requiring only two parameters, it also is comparatively computationally expensive, as previously shown by Rewicki et al.~\cite{rewicki2023is}.
The processing of the unreduced HLT dataset required around 100 h using 16 parallel processes on a high-end server CPU.
Optimising the parameters of the one-liner methods for each of the data channels takes less, but still significant computational effort.
At the same time, all of these methods are offline algorithms, and the competitive results of the one-liner methods was achievable only by optimising their parameters to maximise the test scores.
The employed learning-based methods however, aside from fine-tuning the q parameter of SPOT which serves as a measure to balance false-positive and false-negative detections~\cite{siffer2017anomaly}, were trained solely on data that has no overlap with the test set.
Even given all described advantages of the baseline algorithms, the DeepHYDRA method still manages to outperform them on the HLT datasets.
Additionally, while hyperparameter tuning and optimisation of the applied augmentations are time-consuming, once the models are trained, the combined DeepHYDRA strategy can process the test set in pure streaming fashion, where GPU memory transfer overhead is significant, in around 0.05 s per sample, or around 20 minutes for the whole testing set.
The combined detection approach has also been tested in a purely CPU-based live detection scenario, running on one of the nodes of the HLT cluster, where the pure detection time per sample without considering the overhead for data retrieval is around 0.15 s.
This makes the method able to monitor the whole cluster with negligible added compute overhead, meaning it is able to run on existing HLT hardware.\newline
The results of the conducted computational and memory intensity study show that computationally expensive anomaly detection methods do not necessarily improve the detection outcome.
In fact, the best-performing combined detection models on the HLT dataset, DeepHYDRA-DAGMM and DeepHYDRA-USAD, are two orders of magnitude smaller in footprint than S-MERLIN, comparable to the one-liner method 4.
The transformer-based methods on the other hand do show competitive detection results, but also have comparatively large computational and memory footprints.\newline
These insights will guide our further efforts of developing computationally and memory-efficient anomaly detection approaches.\newline
The detection of point anomalies, global and contextual remains a challenge, and we strive to improve the detection of such anomalies in future work.
Our early results for the modified Eclipse dataset~\cite{aksar2023dataset} have also been mixed thus far.
While T-DBSCAN on its own shows exceptional performance, we pose that a hyperparameter search and an optimization of augmentation parameters, as was done for the HLT datasets, will enable DeepHYDRA to achieve competitive performance on this dataset, and we aim to improve on the presented results in continued work.
Such work should also attempt to increase the detection difficulty of the reshaped Eclipse dataset, which has so far been relatively low as is evident observing our results for the one-liner baseline methods.\newline
A further development step is to evaluate HDBSCAN in place of DBSCAN, which is less sensitive to parameter selection~\cite{mcinnes2017hdbscan}, and may simplify the application to other datasets.
Future work on DeepHYDRA will also include the evaluation of other dimensionality reduction methods in place of T-DBSCAN such as PCA or UMAP, which this work is lacking so far.
While this work included computational intensity and memory footprint evaluation purposefully reported in a datatype-agnostic fashion, possible future work should also study the impact of different datatype sizes as well as methods such as pruning and quantization on overhead and performance.

\section{Conclusion}
In this work, we lay out the problems faced in deep-learning-based anomaly detection in environments in which system configuration is highly-variable, with the ATLAS HLT system serving as a case study.
This problem is addressed through the introduction of the Deep Hybrid DBSCAN/Reduction-Based Anomaly Detection (DeepHYDRA) approach, which combines the advantages of both T-DBSCAN-based detection for precise detection of short-term anomalies with the advantages of deep-learning-based detection to capture long-term anomalies.
Two strategies for improving the performance of the learning-based detection are introduced, consisting of data augmentation and semi-supervised training.
The proposed approach is evaluated on the presented novel HLT datasets and modified Eclipse datasets, with results showing that the T-DBSCAN-based and deep-learning-based detection approaches together offer superior performance on short and long-term anomalies compared to the individual methods and to the non-ML-based baseline algorithms.
The presented data augmentation and semi-supervised training strategies are shown to both individually be able to improve detection performance, while the combined detection strategy followed in the DeepHYDRA method is shown to outperform each individual method as well as the baseline methods.
While the Eclipse dataset remains a challenge for the combined detection approach, the conducted characterisation of computational intensity and memory footprint outlines DeepHYDRA as a computationally inexpensive approach that is able to make learning-based anomaly detection highly scalable and extend it to the area of variably-configured systems.

\paragraph{Acknowledgements}
This work has been sponsored by the Wolfgang Gentner Programme of the German Federal Ministry of Education and Research (grant no. 05E18CHA).

\bibliographystyle{plain}
\bibliography{paper_arxiv}

\end{document}